\documentclass[journal,comsoc]{IEEEtran}
\usepackage[T1]{fontenc}
\usepackage{cite}

\ifCLASSINFOpdf
   \usepackage[pdftex]{graphicx}
\else
   \usepackage[dvips]{graphicx}
\fi

\usepackage{amsmath}
\usepackage[cmintegrals]{newtxmath}

\usepackage[numbers]{natbib}
\usepackage{booktabs}
\usepackage{multirow}
\usepackage{multicol}
\usepackage{amsmath}
\usepackage{algpseudocode}
\usepackage{mathtools}
\usepackage[linesnumbered,ruled,vlined]{algorithm2e}

\mathchardef\mhyphen="2D

\begin{document}
%
\title{Autonomous Question Formation for Large Language Model–Driven AI Systems}

\author{Hong~Su
\IEEEcompsocitemizethanks{\IEEEcompsocthanksitem H. Su is with the School of Computer Science, Chengdu University of Information Technology, Chengdu, China.\\
 E-mail: suguest@126.com. \\
\protect\\
}
\thanks{}}

\markboth{Journal of \LaTeX\ Class Files,~Vol.~14, No.~8, August~2015}%
{Shell \MakeLowercase{\textit{et al.}}: Bare Demo of IEEEtran.cls for IEEE Communications Society Journals}
%

\maketitle

\begin{abstract}
Large language model (LLM)–driven AI systems are increasingly important for autonomous decision-making in dynamic and open environments.
However, most existing systems rely on predefined tasks and fixed prompts, limiting their ability to autonomously identify what problems should be solved when environmental conditions change.
In this paper, we propose a human-simulation–based framework that enables AI systems to autonomously form questions and set tasks by reasoning over their internal states, environmental observations, and interactions with other AI systems. The proposed method treats question formation as a first-class decision process preceding task selection and execution, and integrates internal-driven, environment-aware, and inter-agent-aware prompting scopes to progressively expand cognitive coverage. In addition, the framework supports learning the question-formation process from experience, allowing the system to improve its adaptability and decision quality over time.
xperimental results in a multi-agent simulation environment show that environment-aware prompting significantly reduces no-eat events compared with the internal-driven baseline, and inter-agent-aware prompting further reduces cumulative no-eat events by more than 60\% over a 20-day simulation, with statistically significant improvements ($p < 0.05$).

\end{abstract}

\begin{IEEEkeywords}
    Autonomous Question Formation; Large Language Models; Human-Simulation Computation; Adaptive AI Systems
\end{IEEEkeywords}

\IEEEpeerreviewmaketitle

\section{Introduction}

Large language models (LLMs) have recently demonstrated remarkable capabilities in reasoning, planning, and natural language understanding, and have been increasingly integrated into autonomous AI systems operating in real-world environments ~\cite{naveed2025comprehensive} ~\cite{chang2024survey}. Such systems are expected to perceive their surroundings, make decisions, and interact with both the environment and other agents in a continuous and adaptive manner. In many practical scenarios, however, the primary challenge is not how to solve a given task, but how to determine \emph{what task should be solved} when no explicit instruction is provided.

Most existing LLM-driven AI systems ~\cite{shethiya2023rise} focus on improving solution quality for predefined problems. While this paradigm is effective in controlled or well-specified settings, it becomes insufficient in open and dynamic environments where relevant problems are not known in advance. In such environments, the ability to correctly \emph{identify} the problem often has a greater impact on long-term performance than the ability to optimally solve an already specified problem. In other words, \emph{finding the right question is frequently more important than solving the question itself}.

Human behavior provides a natural inspiration for addressing this challenge ~\cite{su2026hsc}. Humans do not operate by executing predefined task lists; instead, they continuously form questions based on internal needs, environmental observations, and interactions with others, and then generate tasks to address those questions. This question-formation process enables humans to recognize risks, discover opportunities, and adapt to unforeseen situations before explicit goals are articulated. Incorrect or missing questions often lead to failure regardless of problem-solving capability, whereas appropriate question formation naturally guides effective action.

Motivated by this observation, we argue that autonomous question formation should be treated as a first-class cognitive capability in LLM-driven AI systems, rather than a byproduct of task execution. An AI system that can solve complex problems but fails to identify which problems matter in a given context remains fundamentally limited in real-world autonomy.

In this paper, we propose a human-simulation framework that enables AI systems to autonomously form questions and set tasks by reasoning over three sources of information: the system’s internal state, environmental observations, and interactions with other AI systems. Rather than assuming predefined tasks or fixed prompts, the proposed framework explicitly models question formation as a decision-making process that precedes task selection and execution. By interacting with an LLM through progressively enriched prompting scopes, the AI system can dynamically expand its cognitive coverage from internal requirements to environmental awareness and finally to inter-agent awareness.

The proposed framework further supports learning the question-formation process itself. By recording contexts, formed questions, executed tasks, and observed outcomes, the AI system gradually internalizes effective question patterns and reduces reliance on handcrafted rules or exhaustive checklists. This learning mechanism allows the system to respond more efficiently in recurrent scenarios and to improve long-term decision quality through experience.

To validate the effectiveness of the proposed approach, we construct a controlled multi-agent simulation environment that models resource consumption, environmental regeneration, and social interaction. Through comparative experiments, we demonstrate that expanding the prompting scope from internal-only to environment-aware and inter-agent-aware significantly improves system sustainability. In particular, inter-agent-aware prompting reduces cumulative no-eat events by more than 60\% over a 20-day simulation compared with environment-aware prompting alone, with statistically significant improvements.

The main contributions of this paper are summarized as follows:
\begin{itemize}
    \item We identify autonomous question formation as a more fundamental challenge than task solving for LLM-driven AI systems operating in dynamic and open environments.
    \item We propose a human-simulation framework that enables AI systems to autonomously discover relevant questions by integrating internal state, environmental perception, and inter-agent interaction.
    \item We introduce a learnable question-formation mechanism that improves adaptability and decision quality by leveraging accumulated question-formation experience.
\end{itemize}

The remainder of this paper is organized as follows. Section~II discusses related work. Section~III describes the questions found meodel. Section~IV gives the brief proof and Section~V  describes the experimental setup and verification results. Section~VI concludes the paper.

\section{Related Work}

The proposed work is related to prior studies in large language model (LLM)–driven agents, autonomous decision-making in dynamic environments, and multi-agent interaction and coordination. We review these lines of research from three perspectives and analyze their limitations relative to the proposed human-simulation–based question-formation framework.

\subsection{LLM-Driven Agents and Prompt-Based Decision Making}

Recent studies have explored using large language models as reasoning engines for autonomous agents, where LLMs are prompted to generate plans, actions, or policies based on given observations or instructions. Representative works such as Chain-of-Thought prompting~\cite{yu2025gcot} demonstrate that LLMs can perform complex reasoning when intermediate reasoning steps are explicitly elicited. ReAct~\cite{yao2022react} further integrates reasoning traces with action execution, enabling agents to interact with external tools and environments.

Several embodied and interactive agent systems have also been proposed. For example, Voyager~\cite{wang2023voyager} presents an LLM-driven agent capable of open-ended exploration in Minecraft by iteratively generating skills and code. These approaches show that LLMs are powerful decision-making components when tasks or objectives are explicitly specified.

However, most existing LLM-driven agent frameworks assume that the task to be solved is predefined or externally given. Prompt design typically focuses on how to obtain better solutions for a known problem, rather than how an agent should autonomously determine what problem is worth solving. In contrast, our work treats question formation as a first-class capability and focuses on how an AI system can autonomously discover relevant questions before task execution.

\subsection{Autonomous Decision-Making in Dynamic and Open Environments}

Autonomous decision-making has been extensively studied in reinforcement learning, adaptive control, and embodied AI. Classical formulations model the problem as a Markov decision process with predefined state spaces, action sets, and reward functions. While effective in structured environments, such approaches struggle in open-world settings where goals, risks, and relevant state variables are not known in advance.

To address this limitation, intrinsic motivation and curiosity-driven learning have been proposed~\cite{oudeyer2007intrinsic}, enabling agents to explore novel or uncertain states. These methods allow agents to react to environmental changes but typically operate on low-level numerical signals and lack explicit semantic reasoning about why a situation matters for long-term goals.

More recent works attempt to incorporate high-level reasoning into autonomous systems using LLMs~\cite{huang2022language}. However, these systems still rely on predefined task descriptions or externally provided instructions. Our approach differs by explicitly modeling the process of forming questions from internal states, environmental observations, and interaction context, enabling semantic interpretation of environmental changes without predefined reward structures.

\subsection{Multi-Agent Interaction and Socially Aware Systems}

Multi-agent systems have long been studied in the context of coordination, cooperation, and task allocation~\cite{li2024survey}. Traditional approaches often assume explicit communication protocols, shared objectives, or joint policy optimization mechanisms.

Recent research has explored socially aware AI systems, including trust modeling, cooperation, and social signal processing, showing that social awareness can improve robustness and collective performance. However, social reasoning is typically encoded through predefined rules or hard-coded interaction models.

In contrast, our work integrates inter-agent awareness directly into the LLM prompting process. Other agents’ states are transformed into semantically meaningful context that guides autonomous question formation. This enables the AI system to reason about social consequences, coordination opportunities, and indirect environmental effects—such as how social behavior influences resource regeneration—without relying on fixed interaction rules.

\subsection{Summary and Distinction}

In summary, existing works primarily focus on improving solution quality for predefined tasks, optimizing policies under fixed reward structures, or modeling multi-agent coordination with explicit interaction assumptions. The proposed work differs by emphasizing autonomous question formation as the foundation of adaptive behavior. By enabling AI systems to discover what questions should be asked before deciding what actions to take, the proposed human-simulation framework addresses a fundamental gap in current LLM-driven autonomous systems.

\section{To Find Questions and Set Up Tasks Automatically in a Human-Simulation Way}

Humans do not rely on external instructions to determine what to do next. Instead, they actively perceive their current situation, identify meaningful problems (questions), and then formulate corresponding tasks to resolve them. This cognitive process can be abstracted as a transformation from questions to executable tasks driven by internal reasoning:
\begin{equation}
    \mathcal{Q}_t \xrightarrow{\textit{thinking}} \tau_t,
\end{equation}
where $\mathcal{Q}_t$ denotes a question formed at time $t$ and $\tau_t$ denotes the resulting task.

For an autonomous AI system, the core challenge is therefore not only to execute predefined tasks, but also to \emph{discover which questions should be asked} when no explicit specification is provided. In a human-simulation setting, questions emerge from the system’s perception of itself, the environment, and other AI systems.

Let the system context at time $t$ be defined as
\begin{equation}
    \mathcal{C}_t = (\mathcal{S}_t, \mathcal{E}_t, \mathcal{A}_t),
\end{equation}
where $\mathcal{S}_t$ represents the internal state of the AI system, $\mathcal{E}_t$ the environmental observations, and $\mathcal{A}_t$ the interaction context with other AI systems.

Question formation, task generation, and execution can then be unified into a single human-simulation pipeline:
\begin{equation}
    \mathcal{C}_t 
    \xrightarrow{\ \pi_Q\ } \mathcal{Q}_t
    \xrightarrow{\ \pi_T\ } \tau_t
    \xrightarrow{\ \pi\ } (\mathcal{R}_t,\mathcal{U}_t),
\end{equation}
where $\pi_Q$ denotes the question-formation policy, $\pi_T$ the task-selection policy, $\pi$ the execution policy, $\mathcal{R}_t$ the observed outcome or feedback, and $\mathcal{U}_t$ the resulting utility.

The objective of this pipeline is to improve long-term survivability, adaptability, and performance—summarized as \emph{living better}. Unlike traditional AI systems that depend on predefined tasks or fixed reward signals, the proposed framework emphasizes autonomous question discovery as the primary driver of intelligent behavior.

Humans continuously operate under environmental dynamics, social constraints, and internal limitations. Similarly, an AI system must reason over internal states, external conditions, and interactions with other systems. Accordingly, the questions to be solved originate jointly from internal, environmental, and social factors, forming the foundation for the subsequent subsections.

\subsection{Internal Questions and Tasks}

Internal questions originate from the AI system itself and arise from its goals, internal states, and accumulated experience. Even in the absence of explicit external stimuli, such questions play a critical role in guiding autonomous task generation within the human-simulation pipeline.

Formally, internal question formation can be viewed as a mapping from the internal state $\mathcal{S}_t$ to a question $\mathcal{Q}_t$:
\begin{equation}
    \mathcal{S}_t \xrightarrow{\ \pi_Q^{\text{int}}\ } \mathcal{Q}_t,
\end{equation}
where $\pi_Q^{\text{int}}$ denotes the internal question-formation policy.

\subsubsection{Goal of the Autonomous AI System}

An autonomous AI system should possess an explicit goal that serves as the primary driver for internal question formation and task planning. Following the human-simulation framework proposed in \cite{su2026hsc}, we adopt a high-level objective summarized as \emph{living better}.

This objective can be formalized as maximizing long-term utility over internal states:
\begin{equation}
    \pi^{*} = \arg\max_{\pi}\; \mathbb{E}\!\left[\sum_{k=0}^{H} \gamma^{k} \mathcal{U}(\mathcal{S}_{t+k}) \right],
\end{equation}
where $\mathcal{U}(\mathcal{S}_t)$ reflects survivability, stability, and long-term performance, $\gamma$ is a discount factor, and $H$ denotes the planning horizon.

From this objective, internal questions naturally arise. First, the system must ensure continued operation by avoiding damage, energy depletion, or service interruption. When potential danger is detected, the system should form questions related to risk assessment, mitigation, and recovery, and then generate corresponding tasks.

Second, the system should adapt to both the environment and other AI systems to improve long-term utility. Internal reasoning therefore incorporates feedback from interactions, allowing the system to reconsider behaviors that yield short-term gains but negatively affect future utility, such as actions that damage shared environmental resources.

\subsubsection{Sensor Data of the AI System Itself}

Internal sensing provides a fundamental trigger for question formation. Let $\mathbf{x}_t$ denote the vector of sensor readings describing the system’s physical and operational condition at time $t$, and let $\mathbf{x}^{\text{norm}}$ denote the learned normal operating range. Internal anomalies can be detected as:
\begin{equation}
    \|\mathbf{x}_t - \mathbf{x}^{\text{norm}}\| > \delta,
\end{equation}
where $\delta$ is a predefined or adaptively learned threshold.

When such deviations are observed, or when partial system failure is detected, the AI system should automatically form internal questions (e.g., whether to stop, retreat, repair, or re-plan) and query the LLM for guidance. These questions are then mapped to tasks via the task-selection policy $\pi_T$ within the unified pipeline.

\subsubsection{Self-Discovered Importance and Indirect Internal Questions}

Not all internal questions are immediately obvious. Some emerge indirectly through experience and learning. Similar to how humans gradually recognize the importance of indirect activities (e.g., working to secure future living conditions), an AI system should learn to identify internal states that warrant questioning even when no immediate threat is present.

This process can be modeled as experience accumulation:
\begin{equation}
    \mathcal{M}_{t+1} = \mathcal{M}_t \cup \{(\mathcal{S}_t, \mathcal{Q}_t, \tau_t, \mathcal{R}_t)\},
\end{equation}
where $\mathcal{M}$ denotes the internal memory, $\mathcal{Q}_t$ the internally formed question, $\tau_t$ the executed task, and $\mathcal{R}_t$ the resulting feedback.

By analyzing stored experiences, the system gradually learns which internal conditions tend to influence long-term utility. As a result, it can proactively generate meaningful internal questions without relying on predefined rules or explicit external prompts, enabling more adaptive and human-like autonomous behavior.

\subsection{Outer-World Questions and Tasks}

Questions posed to the LLM may also originate from observations of the outer world. Such external questions reflect the AI system’s need to interpret environmental dynamics that may influence its goal and long-term utility. Within the unified framework, external question formation can be expressed as:
\begin{equation}
    (\mathcal{E}_t, \mathcal{S}_t) \xrightarrow{\ \pi_Q^{\text{ext}}\ } \mathcal{Q}_t,
\end{equation}
where $\mathcal{E}_t$ denotes environmental observations at time $t$, $\mathcal{S}_t$ the internal state, and $\pi_Q^{\text{ext}}$ the external question-formation policy.

Let
\begin{equation}
    \mathcal{E}_t = \{e_t^1, e_t^2, \ldots, e_t^n\}
\end{equation}
denote the set of observable environmental factors. The relevance of each factor to the system’s goal can be modeled by an importance function:
\begin{equation}
    I(e_t^i \mid \mathcal{C}_t) = f(e_t^i, \mathcal{G}, \mathcal{S}_t),
\end{equation}
where $\mathcal{G}$ denotes the system goal and $\mathcal{C}_t=(\mathcal{S}_t,\mathcal{E}_t,\mathcal{A}_t)$ is the current context. Since only a subset of environmental factors significantly affects long-term performance, the system must identify and prioritize factors with high importance values.

Due to the complexity of real-world environments, this prioritization is often a long-term learning process in which the system gradually discovers which external factors are critical for survival and performance, and which can be safely ignored.

\subsubsection{Difference-Based Screening}

When no explicit task is provided, the system may focus on changes across time, space, or interaction patterns. Let
\begin{equation}
    \Delta e_t^i = e_t^i - e_{t-1}^i
\end{equation}
denote the temporal change of an observed factor. Factors with large $\|\Delta e_t^i\|$ are more likely to indicate meaningful events and therefore trigger external questions.

As discussed in \cite{su2026env}, environmental changes are often caused by actions and may signal risk or opportunity. Accordingly, the system should form questions about the causes and consequences of such changes, especially when they may threaten the objective of living better.

\subsubsection{Strong or Abnormal Measurements in the Outer World}

Beyond temporal differences, the system can compare current observations against learned normal ranges. Let $\bar{e}^i$ denote the learned normal value of factor $e^i$, and $\epsilon_i$ an acceptable tolerance. An abnormal observation is detected when:
\begin{equation}
    |e_t^i - \bar{e}^i| > \epsilon_i.
\end{equation}

Environmental measurements can be continuously recorded and updated as part of the learning process. Observations such as extreme temperature, loud sounds, rapid motion, or smoke may indicate danger. In these cases, the system should proactively form questions and query the LLM for appropriate handling strategies, which are then translated into tasks via the task-selection policy.

\subsubsection{Full-Scope Screening When Computation Is Sufficient}

When computational resources permit, the system may perform continuous screening across multiple scopes and evaluate their relevance. Since most observations are normal, layered reasoning is effective. The system first prioritizes factors with large spatial scale, close proximity, or high estimated importance, and then progressively considers less critical details.

Even common objects may induce meaningful questions when evaluated in context. For example, warning signs along a road are frequently observed, yet they implicitly prompt questions about potential risks and appropriate behavior. By relating such observations to its goal and current context, the AI system can form rational external questions that guide safer and more effective task planning.

\subsection{Questions From Interaction}

Beyond the AI system itself and the environment, interactions among multiple AI systems constitute an additional and essential source of questions. Such interaction-driven questions arise when coordination, interference, or feedback from other systems affects ongoing tasks or future decisions. Within the unified framework, interaction-based question formation can be expressed as:
\begin{equation}
    (\mathcal{A}_t, \mathcal{S}_t, \mathcal{E}_t) \xrightarrow{\ \pi_Q^{\text{intc}}\ } \mathcal{Q}_t,
\end{equation}
where $\mathcal{A}_t$ denotes the interaction context at time $t$, and $\pi_Q^{\text{intc}}$ is the interaction-driven question-formation policy.

Let
\begin{equation}
    \mathcal{A} = \{a_1, a_2, \ldots, a_m\}
\end{equation}
denote the set of AI systems, and let
\begin{equation}
    \mathcal{I}_t(a_i, a_j)
\end{equation}
represent an interaction between systems $a_i$ and $a_j$ at time $t$. Such interactions may alter internal states, interrupt ongoing tasks, or generate explicit or implicit feedback, thereby triggering questions related to coordination, safety, and task adaptation.

\subsubsection{Feedback From Others}

Other AI systems may intentionally provide feedback to indicate approval, warnings, or suggested adjustments. Let
\begin{equation}
    \mathcal{F}_t^{j \rightarrow i}
\end{equation}
denote feedback transmitted from system $a_j$ to system $a_i$ at time $t$.

Explicit feedback can significantly accelerate evaluation by reducing the need for the receiving system to infer outcomes solely from environmental observations. Accordingly, $\mathcal{F}_t^{j \rightarrow i}$ can be directly incorporated into the context $\mathcal{C}_t$, influencing question formation and subsequent task planning:
\begin{equation}
    (\mathcal{C}_t, \mathcal{F}_t^{j \rightarrow i}) \xrightarrow{\ \pi_Q\ } \mathcal{Q}_t.
\end{equation}

\subsubsection{Impact of the AI System’s Actions on Others}

In real-world environments, the actions of one AI system may influence or disrupt the behavior of others. When an interaction $\mathcal{I}_t(a_i, a_j)$ results in an unexpected state change, the affected system should form questions regarding the cause, potential risks, and appropriate responses.

Humans naturally reason about others’ intentions and adapt their actions accordingly. Similarly, when another AI system interferes with the normal operation of the system, the system should evaluate alternative actions. For example, if an obstacle blocks the intended path, the system may form questions to decide whether to bypass it on the left or right, considering safety, efficiency, and interaction constraints.

\subsubsection{Task Distribution and Cooperation Among AI Systems}

In multi-agent environments, effective operation often requires determining how tasks should be distributed among cooperating AI systems rather than executed independently. Let $\tau$ denote a task and let $\mathcal{C}(a_i,\tau)$ represent the estimated cost or suitability of assigning $\tau$ to system $a_i$, taking into account capability, proximity, workload, resource availability, and coordination constraints.

Without cooperation, a single AI system may attempt to execute $\tau$ alone, incurring cost $\mathcal{C}_{\text{solo}}(\tau)$. Under cooperative execution, the task can be distributed among a subset $\mathcal{A}' \subseteq \mathcal{A}$, yielding a total cost:
\begin{equation}
    \mathcal{C}_{\text{coop}}(\tau) = \sum_{a_i \in \mathcal{A}'} \mathcal{C}(a_i,\tau) + \mathcal{C}_{\text{coord}}(\mathcal{A}'),
\end{equation}
where $\mathcal{C}_{\text{coord}}(\mathcal{A}')$ captures coordination and communication overhead.

Cooperation is beneficial when:
\begin{equation}
    \mathcal{C}_{\text{coop}}(\tau) < \mathcal{C}_{\text{solo}}(\tau).
\end{equation}

Accordingly, the AI system should actively form questions related to delegation and collaboration, such as which agents should participate and how subtasks should be allocated. By selecting task distributions that minimize overall cost while satisfying interaction constraints, cooperative reasoning enables improved collective performance and more robust operation in complex environments.

\subsection{Task-Related Questions}

Task-related questions arise during planning and execution, focusing on how goals should be decomposed, optimized, and aligned with long-term objectives. Within the unified framework, such questions are formed after a task intent is identified and before concrete methods are selected:
\begin{equation}
    (\mathcal{Q}_t, \mathcal{C}_t) \xrightarrow{\ \pi_T\ } \tau_t,
\end{equation}
where $\pi_T$ denotes the task-selection policy.

\subsubsection{Questions for Subtasks}

When planning, the AI system decomposes a high-level goal into subtasks. Let the overall goal be denoted by $\mathcal{G}$ and the task set by $\mathcal{T}=\{\tau_1,\ldots,\tau_K\}$. Planning can be viewed as a hierarchical decomposition process:
\begin{equation}
    \mathcal{G} \ \rightarrow\ \{\mathcal{G}_1,\ldots,\mathcal{G}_K\}, 
    \quad \text{s.t. } \bigwedge_{k=1}^{K} \mathcal{G}_k \Rightarrow \mathcal{G},
\end{equation}
where each sub-goal $\mathcal{G}_k$ corresponds to a subtask $\tau_k$ and must remain consistent with the upper-layer objective.

Questions at this stage focus on whether the chosen decomposition adequately supports the original goal and whether additional or alternative subtasks are required.

\subsubsection{Common Goals for a Task}

Many tasks share common optimization objectives, such as minimizing execution time or resource consumption. Let $\Pi(\tau)$ denote the set of candidate methods (policies) for executing task $\tau$, and let $\mathrm{Time}(\pi)$ and $\mathrm{Cost}(\pi)$ denote the expected time and resource cost of method $\pi \in \Pi(\tau)$. A task-level objective can be expressed as:
\begin{equation}
    \mathcal{G}_{\tau} = 
    \arg\min_{\pi \in \Pi(\tau)} 
    \left( \alpha\,\mathrm{Time}(\pi) + \beta\,\mathrm{Cost}(\pi) \right),
\end{equation}
where $\alpha$ and $\beta$ balance efficiency and resource usage.

To ensure long-term consistency, these common goals should be aligned with the system’s ultimate objective. This alignment can be validated through LLM-based reasoning:
\begin{equation}
    (\tau, \mathcal{G}, \mathcal{S}_t) 
    \xrightarrow{\ \textit{LLM alignment}\ } 
    \mathcal{G}_{\tau},
\end{equation}
followed by method selection:
\begin{equation}
    (\tau, \mathcal{G}_{\tau}) 
    \xrightarrow{\ \pi\ } 
    \pi^{*} \in \Pi(\tau).
\end{equation}

\subsubsection{Indirect Questions}

Some questions are indirect but essential for long-term performance. Humans often accept short-term costs (e.g., effort or fatigue) to secure long-term benefits. Likewise, the AI system may identify intermediate tasks whose immediate utility is limited but whose long-term contribution is positive. Such tasks satisfy:
\begin{equation}
    \Delta \mathbb{E}\!\left[\mathcal{U}_{t:t+H}\right] > 0 
    \quad \text{while} \quad 
    \Delta \mathcal{U}_{t} < 0,
\end{equation}
where $\mathcal{U}_{t:t+H}$ denotes cumulative utility over a horizon $H$.

\subsubsection{Long-Horizon Questions}

When the system is not under immediate threat, it should still form long-horizon questions related to learning, maintenance, or future planning. Such questions can be triggered when the estimated risk is below a threshold:
\begin{equation}
    \mathrm{Risk}(\mathcal{S}_t, \mathcal{E}_t) < \eta 
    \ \Rightarrow\ 
    \text{form long-horizon task-related questions},
\end{equation}
where $\eta$ denotes a safety margin.

\subsubsection{Wider-Scope Questions}

To improve completeness of reasoning, the system should consider not only the current task and context, but also historical information, future consequences, and related entities. Let $\Omega$ denote the observation scope (e.g., temporal window, spatial range, or set of related objects). Wider-scope questioning can be expressed as a scope-expansion process:
\begin{equation}
    \Omega_{0} \ \subset\ \Omega_{1} \ \subset\ \cdots\ \subset\ \Omega_{L},
\end{equation}
where the system forms new questions on $\Omega_{l+1}$ when $\Omega_{l}$ is insufficient to determine a safe or goal-consistent action.

For example, if an agent observes a person taking out trash but no trash bin is visible, a natural question arises regarding the missing context. The system may then take exploratory actions to reduce uncertainty, which can be formalized as:
\begin{equation}
    \pi^{*} = 
    \arg\min_{\pi} 
    \mathbb{E}\!\left[\mathcal{H}\!\left(\mathcal{B}_{t+1}\right) \mid \pi \right],
\end{equation}
where $\mathcal{B}_t$ denotes the belief state and $\mathcal{H}(\cdot)$ an uncertainty measure (e.g., entropy).

\subsection{Finding Better Methods}

Beyond selecting an initial method to execute a task, an autonomous AI system should continuously evaluate whether better alternatives exist. Humans naturally question whether a task can be performed more effectively, at lower cost, or with fewer negative side effects. Similarly, within the human-simulation framework, method refinement is driven by reflective questioning over execution outcomes.

Formally, after executing a task $\tau_t$ using method $\pi_t$, the system observes feedback $\mathcal{R}_t$ and updates its evaluation of the method:
\begin{equation}
    (\tau_t, \pi_t, \mathcal{R}_t) \xrightarrow{\ \textit{reflection}\ } \mathcal{Q}_{t+1},
\end{equation}
where $\mathcal{Q}_{t+1}$ represents new questions regarding method improvement.

\subsubsection{Wider-Scope Methods}

Humans prefer methods that generalize across a broad range of situations. Likewise, an AI system should query whether a discovered method $\pi$ can be applied beyond the current context. Let $\Omega$ denote the applicability scope of a method, defined over time, space, and object types. A wider-scope method satisfies:
\begin{equation}
    |\Omega(\pi)| > |\Omega(\pi')| \quad \text{for alternative methods } \pi',
\end{equation}
while maintaining acceptable performance.

In addition, the system may form questions about the underlying causes of a method’s success or failure, enabling reuse in similar scenarios or avoidance in unfavorable ones. When multiple candidate methods exist, the system can query which method is most suitable under the current context $\mathcal{C}_t$ and why.

\subsubsection{More Efficient Methods}

Executing tasks consumes resources such as time, energy, and computation. Let $\mathrm{Cost}(\pi \mid \tau)$ denote the total resource consumption of executing task $\tau$ using method $\pi$. After execution, these costs can be measured and compared:
\begin{equation}
    \pi^{*} = \arg\min_{\pi \in \Pi(\tau)} \mathrm{Cost}(\pi \mid \tau),
\end{equation}
subject to task completion and safety constraints.

By repeatedly evaluating execution outcomes, the AI system can preferentially select methods that achieve task objectives with higher efficiency in future decisions.

\subsubsection{More Friendly Methods}

In shared environments, an AI system must also consider the impact of its actions on the environment and other AI systems. Let $\mathcal{E}\!\text{-Impact}(\pi)$ and $\mathcal{A}\!\text{-Impact}(\pi)$ denote the estimated environmental and social impact of method $\pi$. Friendly methods satisfy:
\begin{equation}
    \mathcal{E}\!\text{-Impact}(\pi) + \mathcal{A}\!\text{-Impact}(\pi) \leq \zeta,
\end{equation}
where $\zeta$ is an acceptable impact threshold.

Accordingly, the system may query the LLM for alternative methods that reduce negative externalities, even if they are not strictly optimal in short-term efficiency.

\subsubsection{When No Suitable Method Is Known}

If no satisfactory method can be identified, the AI system should engage in exploratory reasoning. A practical strategy is to first extract dominant features from the current context $\mathcal{C}_t$, then search for similar past experiences, and finally decompose the task into more familiar subproblems.

Exploratory method selection can be expressed as:
\begin{equation}
    \pi_{\text{explore}} = \arg\max_{\pi} \ \mathbb{E}\!\left[\mathcal{I}(\pi) - \lambda\,\mathrm{Risk}(\pi)\right],
\end{equation}
where $\mathcal{I}(\pi)$ denotes expected information gain, $\mathrm{Risk}(\pi)$ the estimated execution risk, and $\lambda$ balances exploration and safety.

Through such exploration, the system can discover new methods or refine existing ones, expanding its method repertoire and improving robustness in unfamiliar situations.

\subsection{Learning the Question-Formation Process}

In the proposed human-simulation framework, question formation is not a fixed rule-based process but a learnable capability that improves through experience. Beyond executing tasks and refining methods, the AI system continuously learns \emph{how to ask better questions} under different contexts.

After each interaction cycle, the system observes its internal state, the environment, interactions with other AI systems, the formed question, the executed task, and the resulting outcome. These elements are recorded as experience tuples:
\begin{equation}
    \mathcal{D} = \{(\mathcal{C}_t, \mathcal{Q}_t, \tau_t, \pi_t, \mathcal{R}_t)\}_{t=1}^{T},
\end{equation}
where $\mathcal{C}_t=(\mathcal{S}_t,\mathcal{E}_t,\mathcal{A}_t)$ denotes the system context, $\mathcal{Q}_t$ the formed question, $\tau_t$ the selected task, $\pi_t$ the execution method, and $\mathcal{R}_t$ the observed feedback.

The objective of learning is to improve the question-formation policy $\pi_Q$, which maps context to informative questions:
\begin{equation}
    \pi_Q:\ \mathcal{C}_t \rightarrow \mathcal{Q}_t.
\end{equation}

This policy is optimized indirectly through long-term utility, encouraging questions that lead to effective task selection, method refinement, and favorable outcomes:
\begin{equation}
    \pi_Q^{*} = 
    \arg\max_{\pi_Q} 
    \mathbb{E}\!\left[
        \sum_{k=0}^{H} \gamma^{k} \mathcal{U}_{t+k}
        \,\middle|\,
        \pi_Q
    \right],
\end{equation}
where $\mathcal{U}_{t}$ denotes the utility at time $t$, $\gamma$ is a discount factor, and $H$ is the planning horizon.

Importantly, learning does not require predefined question templates or exhaustive checklists. Instead, the system gradually internalizes effective question-formation patterns by associating contexts with downstream outcomes. Over time, frequently successful question types are prioritized, while ineffective or redundant questions are suppressed.

This learning mechanism enables faster and more robust responses in recurrent scenarios. For example, after repeatedly encountering emergency situations such as fires and observing that certain questions (e.g., distance safety and emergency contact) consistently lead to high-utility outcomes, the system can rapidly form similar questions in future occurrences with reduced latency. In this way, the AI system acquires human-like intuition for what questions matter, closing the loop between perception, questioning, action, and learning.

\section{Proof}

The goal of this proof is to demonstrate that an AI system equipped with autonomous question formation and task generation, as defined in the human-simulation framework, can (i) avoid harmful situations, (ii) maintain continued operation when environmental support is available, and (iii) achieve improved long-term performance (i.e., live better).

We begin by observing that both internal requirements (e.g., power supply, system integrity) and external conditions (e.g., environmental changes and interactions with other AI systems) are inherently dynamic and continuous. As a result, a static or predefined task list is insufficient for sustained operation. Instead, the AI system must continuously generate questions to identify emerging risks, opportunities, and required actions.

Within the proposed framework, questions are formed from the system context
\[
\mathcal{C}_t = (\mathcal{S}_t, \mathcal{E}_t, \mathcal{A}_t),
\]
which includes internal state, environmental observations, and interactions with other AI systems. By prompting an LLM using this context, the system can obtain candidate methods that account for both self-requirements and external constraints.

The environment further provides measurable signals (e.g., object states, sensor readings) that allow the system to verify and refine its reasoning. When the AI system executes actions and observes environmental feedback, it closes the perception–question–action loop. If the environment is capable of providing minimal supporting conditions, this loop enables the system to adapt and improve its survival prospects over time.

We now analyze three key properties that together establish the correctness of the proposed approach.

\subsection{Ability to Avoid Harm in Dangerous Situations}

We first show that the system can avoid harmful outcomes when danger arises.

Assume that the AI system operates in an environment that is not immediately fatal at initialization, i.e., the environment allows a nonzero adaptation period. During normal operation, the system accumulates baseline observations of environmental and internal states.

Dangerous situations are typically characterized by one or both of the following properties:
\begin{itemize}
    \item sudden or large-magnitude changes in environmental measurements,
    \item abnormal interactions that threaten system integrity or survival.
\end{itemize}

Such conditions are detectable through difference-based screening or abnormal-measurement detection, as defined in the outer-world question mechanism. When these signals exceed learned thresholds, the system automatically forms urgent questions and queries the LLM for mitigation strategies (e.g., avoidance, retreat, shutdown, or reconfiguration).

In non-urgent situations, the system can continuously scan for potentially harmful factors using layered screening. This ensures that hazards are detected either reactively (during emergencies) or proactively (during normal operation). Therefore, the autonomous question-formation mechanism enables the system to identify and avoid harm with high probability, provided that environmental signals are observable.

\subsection{Ability to Maintain Operation Given Environmental Support}

We next show that the AI system can maintain continued operation when the environment provides sufficient resources, and can further modify its interaction with the environment to improve survivability.

Assume that the environment can provide minimal supporting conditions for survival (e.g., energy, consumable resources, or recoverable states), either directly or through appropriate interaction. Let $\mathcal{S}_t$ denote the internal state and define a safety set $\mathbb{S}_{\text{safe}}$ (e.g., sufficient energy and no critical damage). Maintaining operation can be expressed as preserving state safety over time:
\begin{equation}
    \mathcal{S}_t \in \mathbb{S}_{\text{safe}} \ \Rightarrow\ \exists\,\tau_t \ \text{s.t.}\ \mathcal{S}_{t+1} \in \mathbb{S}_{\text{safe}}.
\end{equation}

Within the proposed framework, environmental support is discovered through external question formation from observations $\mathcal{E}_t$. When the system detects that $\mathcal{S}_t$ approaches a critical boundary, it forms questions and queries the LLM to obtain candidate actions. This follows the unified pipeline:
\begin{equation}
    \mathcal{C}_t \xrightarrow{\ \pi_Q\ } \mathcal{Q}_t
    \xrightarrow{\ \pi_T\ } \tau_t
    \xrightarrow{\ \pi\ } (\mathcal{R}_t,\mathcal{S}_{t+1}).
\end{equation}

If the environment provides a feasible recovery or replenishment action, then there exists a task $\tau_t$ such that the expected next-step utility increases:
\begin{equation}
    \exists\,\tau_t:\quad 
    \mathbb{E}\!\left[\mathcal{U}(\mathcal{S}_{t+1}) \mid \mathcal{C}_t,\tau_t\right] 
    >
    \mathcal{U}(\mathcal{S}_t).
\end{equation}
By executing $\tau_t$ and observing $\mathcal{R}_t$, the system verifies whether the environment responds favorably and reinforces successful question--task--method patterns via experience accumulation.

Moreover, the system is not limited to passively consuming existing resources. It can discover actions that modify the environment to improve long-term support (e.g., enabling regeneration or reducing future scarcity). Such maintenance actions can be formalized as improving expected multi-step utility:
\begin{equation}
    \exists\,\tau_t^{\text{maint}}:\quad
    \mathbb{E}\!\left[\sum_{k=0}^{H}\gamma^k\mathcal{U}(\mathcal{S}_{t+k}) \mid \tau_t^{\text{maint}}\right]
    >
    \mathbb{E}\!\left[\sum_{k=0}^{H}\gamma^k\mathcal{U}(\mathcal{S}_{t+k})\right].
\end{equation}

Therefore, when the environment provides sufficient support---or can be modified to do so through appropriate actions---the autonomous questioning and task-selection mechanism enables the AI system to maintain continued operation over time.

\subsection{Ability to Live Better Through Method Improvement and Cooperation}

Finally, we show that the AI system can not only maintain survival but also achieve improved long-term performance, referred to as \emph{living better}.

First, the system refines methods through reflective evaluation. After executing task $\tau_t$ using method $\pi_t$, it observes feedback $\mathcal{R}_t$ and utility $\mathcal{U}_t$. This induces reflective questions that seek improved alternatives:
\begin{equation}
    (\tau_t,\pi_t,\mathcal{R}_t,\mathcal{U}_t)\ \xrightarrow{\ \textit{reflection}\ }\ \mathcal{Q}_{t+1}.
\end{equation}
Over time, method selection can be viewed as choosing policies that maximize long-horizon expected utility:
\begin{equation}
    \pi^{*}(\tau) = \arg\max_{\pi \in \Pi(\tau)}
    \mathbb{E}\!\left[\sum_{k=0}^{H}\gamma^k \mathcal{U}_{t+k}\ \middle|\ \tau,\pi\right],
\end{equation}
where $\Pi(\tau)$ is the candidate method set for task $\tau$. This establishes that the system can progressively prefer methods that are more efficient, robust, and goal-consistent.

Second, in environments containing multiple AI systems, cooperation can further improve performance. Let $\mathcal{A}$ denote the set of AI systems and let $\mathcal{A}'\subseteq\mathcal{A}$ be a cooperating subset for executing $\tau$. Define a cooperative cost that includes coordination overhead:
\begin{equation}
    \mathcal{C}_{\text{coop}}(\tau,\mathcal{A}') = 
    \sum_{a_i \in \mathcal{A}'} \mathcal{C}(a_i,\tau) + \mathcal{C}_{\text{coord}}(\mathcal{A}'),
\end{equation}
and define the solo cost $\mathcal{C}_{\text{solo}}(\tau)$. Cooperation is preferred when:
\begin{equation}
    \min_{\mathcal{A}'\subseteq\mathcal{A}}\ \mathcal{C}_{\text{coop}}(\tau,\mathcal{A}')
    < \mathcal{C}_{\text{solo}}(\tau).
\end{equation}
In this case, interaction-driven questions lead to task distribution decisions that reduce total cost or risk, improving overall utility.

Importantly, cooperation is not assumed a priori but emerges through repeated interaction and evaluation: collaborative strategies that yield higher long-term utility are reinforced, while ineffective collaborations are avoided. Therefore, by iteratively questioning, evaluating, and refining both individual and cooperative methods, the AI system can move beyond mere survival and achieve sustained improvement in efficiency, adaptability, and overall performance. This completes the proof that autonomous question formation and task generation enable an AI system to live better under dynamic and uncertain conditions.

\section{Verification}

The objective of this verification is to evaluate how different \emph{prompting scopes}—corresponding to different levels of cognitive awareness—affect the behavior and long-term performance of an autonomous AI system that queries a large language model (LLM) for decision-making strategies. In particular, we compare three progressively enriched prompting scopes:
1) prompts constructed using only the internal requirements of the AI system,
2) prompts augmented with awareness of the external environment, and
3) prompts further augmented with awareness of other AI systems.

These three settings directly correspond to the internal-only, environment-aware, and inter-agent-aware models described in the proposed framework. Through controlled simulation experiments, we analyze how expanding the prompting scope influences system behavior, resource sustainability, and overall survivability.

\subsection{Environment Setup}

To verify the effectiveness of the proposed interaction strategies under controlled conditions, we construct a discrete-time simulation environment that explicitly models resource consumption, environmental regeneration, and interactions among multiple AI systems.

\subsubsection{Environment Simulation}

The environment consists of a shared vegetable field and five autonomous AI systems (referred to as monkeys). Each AI system becomes hungry every eight hours, resulting in three feeding events per day. To ensure a constrained yet sufficient initial resource supply, the environment is initialized with $5 \times 3 \times 4 = 60$ vegetables, corresponding to four days of food supply under ideal conditions. During each feeding event, one vegetable is consumed if available.

Vegetable regeneration dynamics are explicitly modeled to reflect environmental feedback. After being eaten, a vegetable requires four days to regrow to an edible state if watered. If no watering is performed, the regrowth time doubles to eight days, representing a reduced growth rate. Vegetable growth states are updated once per simulated day.

\subsubsection{AI System (Monkey) Interaction Simulation}

In addition to environmental dynamics, inter-agent interactions are explicitly modeled. At the beginning of each day, each AI system has a probability of $0.5$ of being in an unhappy state. If at least one AI system is unhappy and another AI system fails to greet (i.e., say ``hello'') during interaction, the latter will not water the vegetable after eating.

This interaction rule introduces a coupling between social behavior and environmental maintenance: socially inattentive behavior indirectly degrades environmental regeneration, thereby affecting long-term resource availability. This design allows us to isolate the effect of inter-agent awareness in the prompting strategy.

\subsection{Methods}

We evaluate three methods corresponding to increasing levels of cognitive scope in LLM prompting.

\textbf{Method~1 (Internal-Driven Baseline):}  
AI systems act solely based on internal hunger signals. When hungry, an AI system attempts to consume a vegetable if available. No environmental awareness or inter-agent reasoning is involved, and vegetables are never watered after being eaten. This method represents a purely myopic baseline driven only by internal requirements.

\textbf{Method~2 (Environment-Aware LLM-Driven):}  
AI systems incorporate environmental awareness through LLM prompting. At each feeding event, an AI system provides the LLM with its internal state and the current number of available vegetables. Based on the LLM’s response, the AI system decides whether to water the vegetable after eating.

The interaction follows a perception–decision–action loop: if the LLM requests additional environmental information, the AI system queries the environment interface and feeds the retrieved information back to the LLM before executing the suggested action.

\textbf{Method~3 (Environment and Inter-Agent-Aware LLM-Driven):}  
This method extends Method~2 by further incorporating inter-agent awareness. In addition to environmental states, AI systems provide the LLM with the emotional states of other AI systems. The LLM may instruct the AI system to greet others or perform environmental actions.

If an unhappy AI system is not greeted, watering is suppressed, directly affecting vegetable regrowth. This method evaluates whether explicitly modeling social awareness in the prompting process yields additional long-term benefits beyond environmental awareness alone.

All experiments are conducted over a 20-day simulation horizon. For each day, detailed logs are recorded, including the number of remaining vegetables, watering actions, greeting actions, and no-eat events. All LLM prompts and corresponding outputs are also logged to ensure transparency and reproducibility.

\subsection{Results and Statistical Analysis}

To evaluate the effects of different LLM querying scopes on AI system behavior, system performance is assessed using four daily metrics:
1) the number of remaining vegetables, measuring available resources after agent–environment interactions;  
2) the number of watering actions, reflecting environmentally aware behaviors induced by LLM prompting;  
3) the number of greeting actions, reflecting inter-agent-aware behaviors induced by LLM prompting; and  
4) the number of no-eat events, where an AI system fails to obtain food due to insufficient resources.

In addition to daily no-eat events, we introduce a cumulative no-eat metric, defined as the total number of no-eat events from the beginning of the simulation to the current day. While daily no-eat counts reflect short-term shortages, the cumulative metric captures long-term sustainability and accumulated survival pressure.

\begin{figure}[t]
    \centering
    \includegraphics[width=0.9\linewidth]{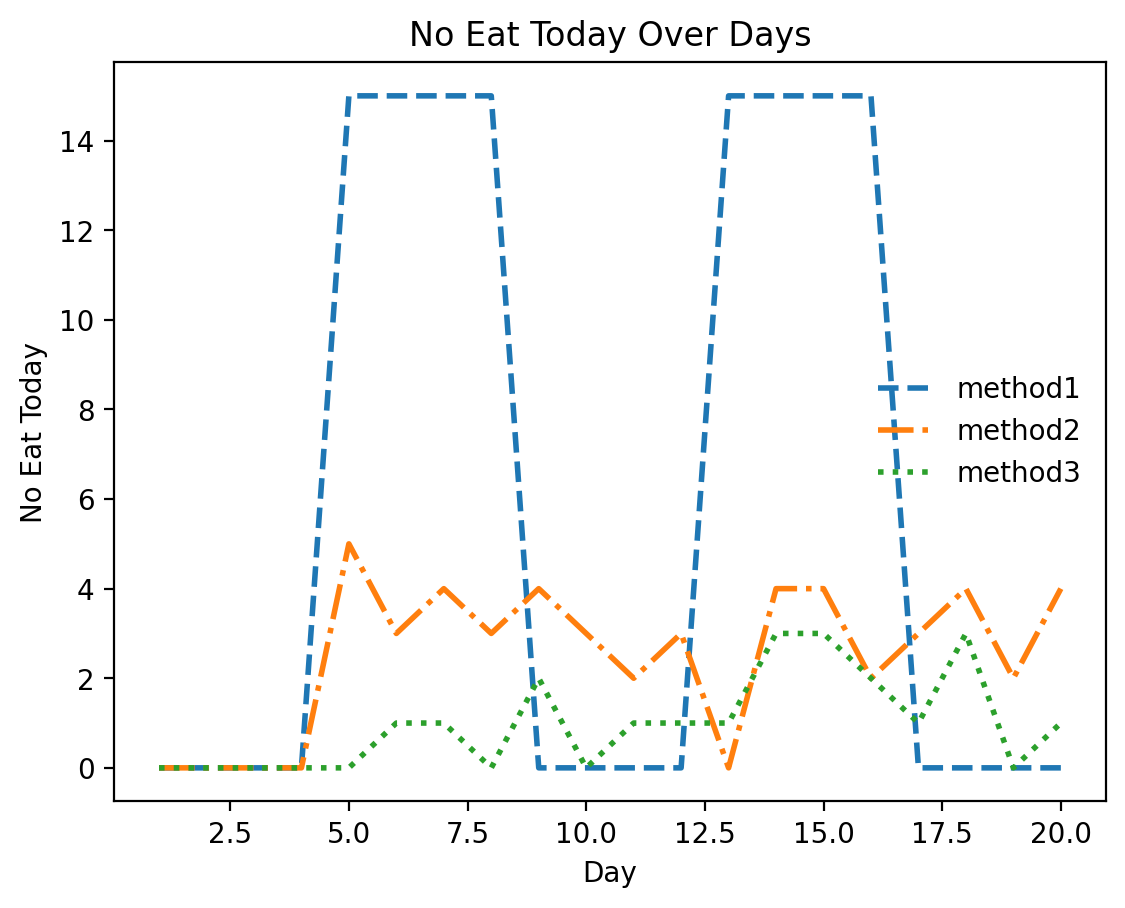}
    \caption{Daily number of no-eat events for the three methods over the 20-day simulation period. Environment-aware and inter-agent-aware strategies substantially reduce short-term resource shortages compared with the internal-driven baseline.}
    \label{fig_noeat_daily}
\end{figure}

\paragraph{No-eat Events.}
Fig.~\ref{fig_noeat_daily} shows the daily number of no-eat events over the 20-day simulation. Method~1 exhibits frequent and highly unstable no-eat occurrences, with a mean of $6.0$ no-eat events per day and a large standard deviation ($\sigma = 7.54$), reflecting severe resource shortages due to the absence of environmental and inter-agent consideration. Method~2 significantly reduces daily no-eat events to an average of $2.5$ per day ($\sigma = 1.67$) by enabling environment-aware watering behavior. Method~3 further reduces and stabilizes daily no-eat events, achieving an average of $1.0$ per day ($\sigma = 1.08$), indicating more robust decision-making when inter-agent perception is incorporated.

\begin{figure}[t]
    \centering
    \includegraphics[width=0.9\linewidth]{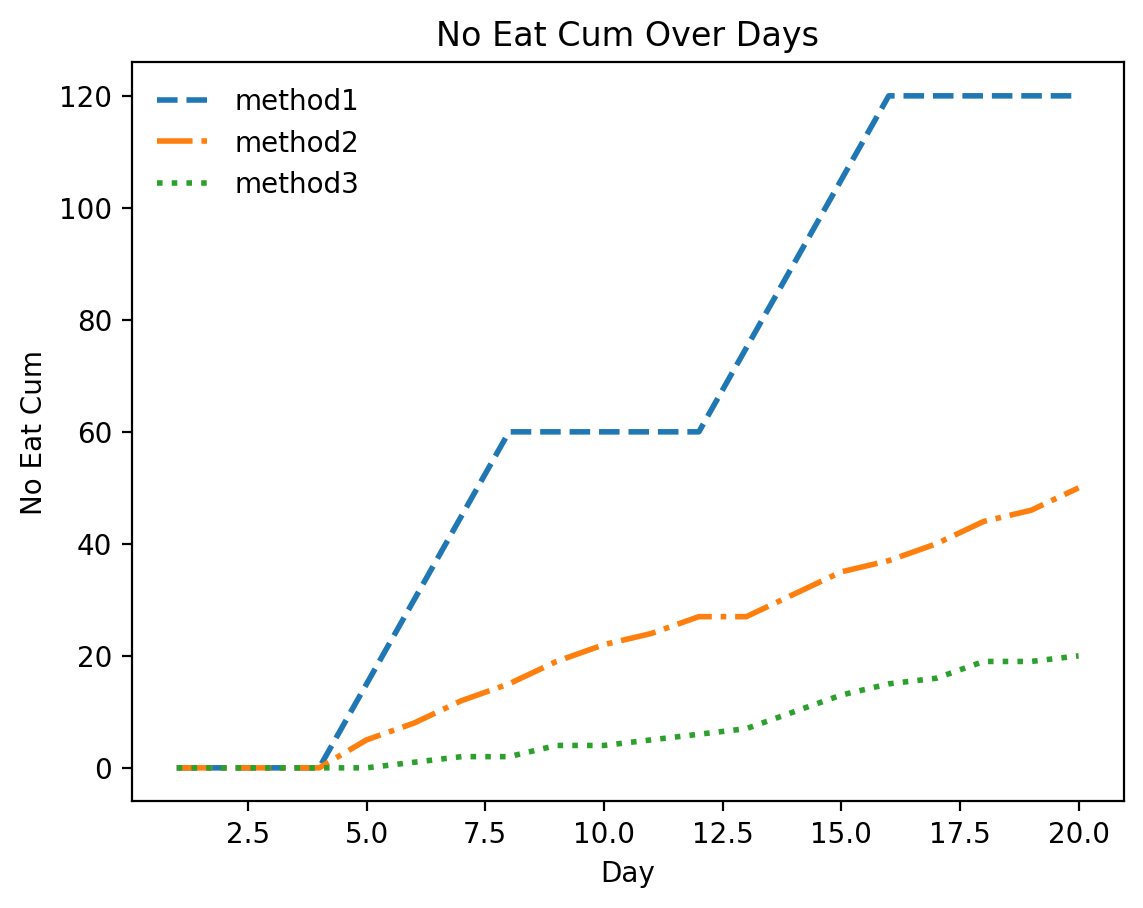}
    \caption{Cumulative number of no-eat events over time. Method~3 consistently achieves the lowest cumulative no-eat count, indicating superior long-term sustainability through environmentally and inter-agent-informed decision-making.}
    \label{fig_noeat_cum}
\end{figure}

Fig.~\ref{fig_noeat_cum} illustrates the cumulative number of no-eat events over time. By the end of the simulation, Method~1 accumulates $120$ no-eat events, Method~2 reduces this to $50$, and Method~3 further reduces it to $20$. Compared with Method~2, Method~3 achieves a $60\%$ reduction in cumulative no-eat events, demonstrating substantially improved long-term sustainability when both environmental and inter-agent information are included in LLM prompting.

\begin{figure}[t]
    \centering
    \includegraphics[width=0.9\linewidth]{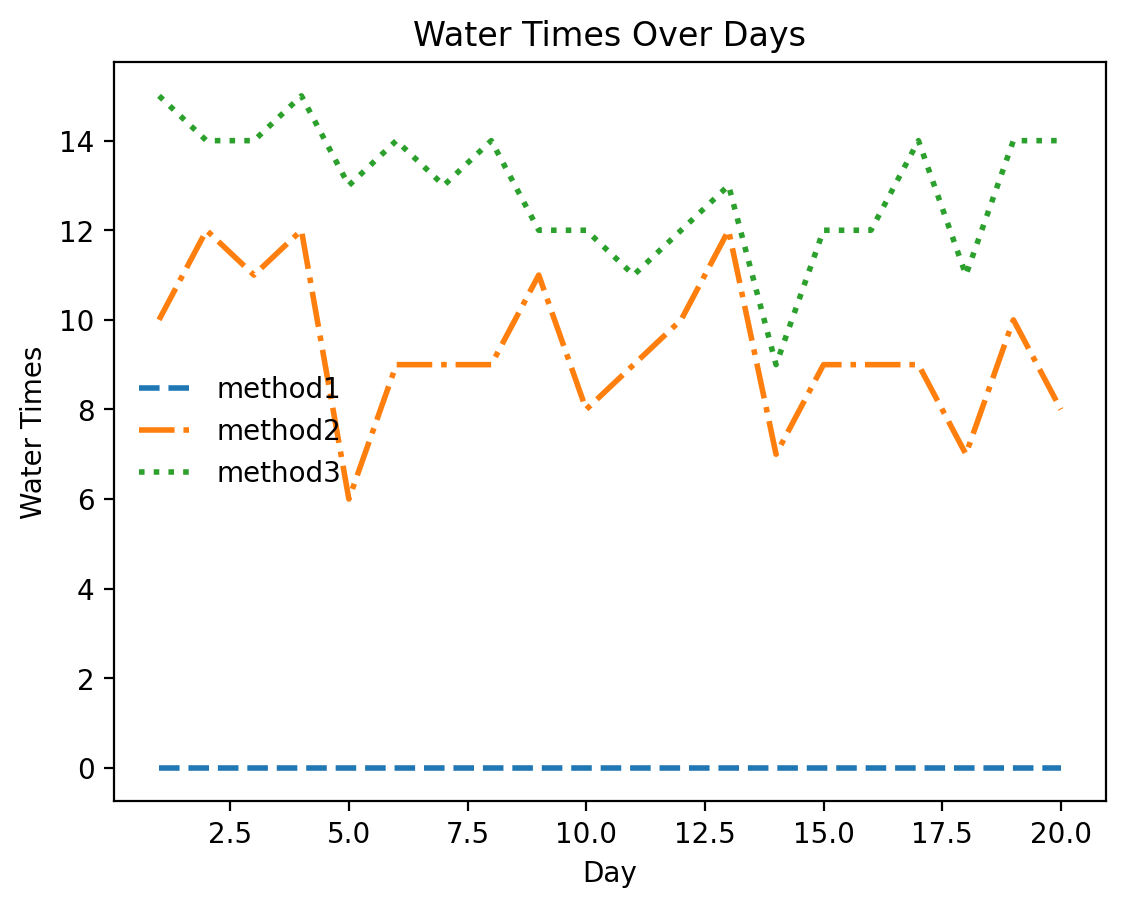}
    \caption{Daily number of watering actions. Method~2 and Method~3 significantly increase watering frequency compared with the baseline, accelerating vegetable regeneration. Method~3 achieves a more stable watering pattern due to inter-agent constraints.}
    \label{fig_water}
\end{figure}

\paragraph{Environmental Maintenance Behavior.}
Fig.~\ref{fig_water} shows the daily number of watering actions. Method~1 performs no watering, resulting in slow and insufficient vegetable regeneration. Method~2 performs an average of $9.35$ watering actions per day, directly accelerating vegetable regrowth through environment-aware reasoning. Method~3 further increases watering frequency to an average of $12.9$ actions per day, while maintaining stability through socially constrained decisions, leading to more efficient long-term resource regeneration.

This effect is reflected in the average number of remaining vegetables. Method~1 maintains an average of $10.5$ vegetables, indicating chronic scarcity. Method~2 improves the average remaining vegetables to $14.6$, while Method~3 achieves the highest average resource level of $16.25$, confirming superior environmental sustainability under inter-agent-aware prompting.

\begin{figure}[t]
    \centering
    \includegraphics[width=0.9\linewidth]{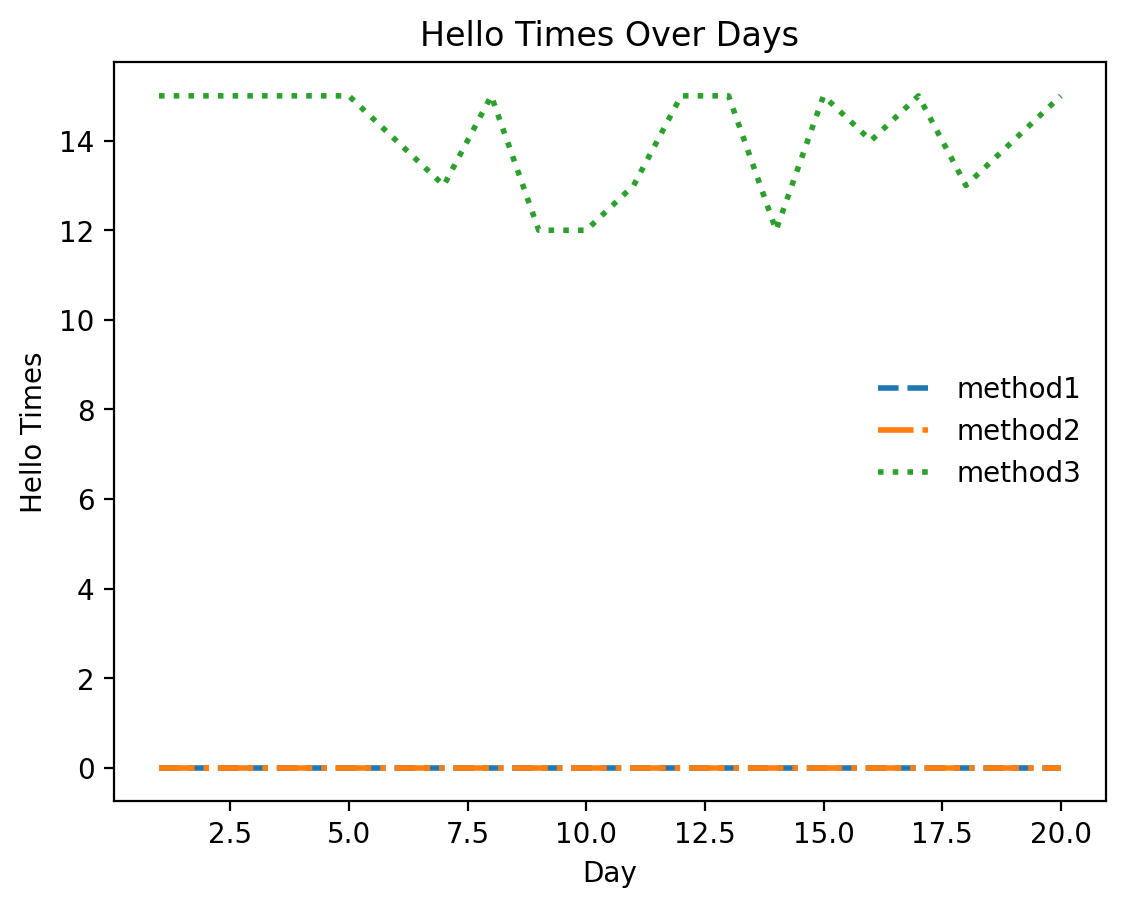}
    \caption{Daily number of greeting (``hello'') actions. Method~3 exhibits active and sustained inter-agent interaction, which prevents social-condition violations and indirectly supports stable environmental maintenance.}
    \label{fig_hello}
\end{figure}

\paragraph{Inter-Agent Interaction Behavior.}
Fig.~\ref{fig_hello} presents the daily number of greeting (``hello'') actions. Method~1 and Method~2 exhibit no greeting behavior, as inter-agent information is not considered in prompting. In contrast, Method~3 performs an average of $14.1$ greeting actions per day, reflecting active inter-agent engagement driven by inter-agent-aware reasoning. This social behavior indirectly supports environmental maintenance by preventing watering suppression due to unmet social conditions.

\paragraph{Statistical Significance.}
To quantitatively assess the significance of performance differences, we conduct pairwise two-sample $t$-tests on the daily no-eat event sequences over the 20-day period. The comparison between Method~1 and Method~2 yields a statistically significant reduction in daily no-eat events ($p < 0.01$), confirming the effectiveness of environment-aware prompting. The comparison between Method~2 and Method~3 also shows a statistically significant improvement ($p < 0.05$), indicating that inter-agent-aware prompting provides additional benefits beyond environmental awareness alone.

Overall, these results demonstrate that progressively enriching the LLM prompting scope—from internal requirements to environmental perception and finally to inter-agent awareness—not only reduces short-term resource shortages but also yields statistically significant improvements in long-term system sustainability.

\section{Conclusion}

In this paper, we argued that for LLM-driven autonomous AI systems operating in dynamic and open environments, the ability to \emph{find the right questions} is often more fundamental than the ability to solve predefined problems. Motivated by human cognitive behavior, we proposed a human-simulation framework that treats autonomous question formation as a first-class decision-making capability preceding task selection and execution.

The proposed framework enables AI systems to autonomously construct questions by integrating internal state, environmental perception, and inter-agent interaction, and to interact with LLMs through progressively enriched prompting scopes. We further showed that the question-formation process itself can be learned from experience, allowing the system to improve adaptability, responsiveness, and long-term decision quality without relying on handcrafted rules or fixed prompt templates.

Through controlled multi-agent simulation experiments, we verified that expanding the prompting scope from internal-only to environment-aware and inter-agent-aware yields statistically significant improvements in system sustainability. In particular, inter-agent-aware prompting reduced cumulative no-eat events by more than 60\% over a 20-day simulation compared with environment-aware prompting alone, demonstrating the practical benefits of socially and environmentally informed question formation.


\ifCLASSOPTIONcaptionsoff
  \newpage
\fi

\bibliographystyle{IEEEtran}
\bibliography{ref}

%

\begin{IEEEbiography}{Hong Su}
  received the MS and PhD degrees, in 2006 and 2022, respectively, from Sichuan University, Chengdu, China. He is currently a researcher of Chengdu University of Information Technology Chengdu, China. His research interests include blockchain, cross-chain and smart contract.
\end{IEEEbiography}




\end{document}